\documentclass{article} 
\usepackage[preprint]{colm2025_conference}

\usepackage{microtype}
\usepackage{hyperref}
\usepackage{url}
\usepackage{booktabs}
\usepackage{graphicx}
\usepackage{algorithm} 
\usepackage{algpseudocode} 
\usepackage{multirow}
\usepackage{amsmath} 
\usepackage{amsfonts} 
\usepackage{amssymb} 
\usepackage{amsthm}
\newtheorem{definition}{Definition}
\newtheorem{theorem}{Theorem}

\usepackage{lineno}

\definecolor{darkblue}{rgb}{0, 0, 0.5}
\hypersetup{colorlinks=true, citecolor=darkblue, linkcolor=darkblue, urlcolor=darkblue}

\title{An Optimization Algorithm for Multimodal Data Alignment}

\author{
  Wei Zhang \\ Department of Electrical Engineering \\ Fudan University \\ \texttt{weizhang123@fudan.edu.cn}
  \And
  Xinyue Wang \\ School of Information Science \\ Fudan University \\ \texttt{xinyue.wang@fudan.edu.cn}
  \And
  Lan Yu \\ Department of Software Engineering \\ Fudan University \\ \texttt{lanyu@fudan.edu.cn}
  \AND
  Shi Li \\ Department of Computer Science \\ Columbia University \\ \texttt{shili081100@columbia.edu}
}

\begin{document}

\ifcolmsubmission
\linenumbers
\fi

\maketitle

\begin{abstract}
In the data era, the integration of multiple data types, known as multimodality, has become a key area of interest in the research community. This interest is driven by the goal to develop cutting-edge multimodal models capable of serving as adaptable reasoning engines across a wide range of modalities and domains. Despite the fervent development efforts, the challenge of optimally representing different forms of data within a single unified latent space—a crucial step for enabling effective multimodal reasoning—has not been fully addressed. To bridge this gap, we introduce AlignXpert, an optimization algorithm inspired by Kernel CCA crafted to maximize the similarities between $N$ modalities while imposing some other constraints. This work demonstrates the impact on improving data representation for a variety of reasoning tasks, such as retrieval and classification, underlining the pivotal importance of data representation. 

\end{abstract}

\section{Introduction}

The advent of the data era has ushered in unprecedented opportunities and challenges in the field of data science and artificial intelligence. Among these, the concept of multimodality, which refers to the integration of multiple types of data, stands out as a critical area of exploration \citep{jewitt2016introducing}. The growing interest in multimodality within the research community is propelled by the arms race to create advanced state of the art multimodal models. These models aim to function as versatile reasoning engines, capable of navigating and interpreting a diverse array of modalities and domains mimicking the human experience. 

However, despite the enthusiastic pursuit of advanced multimodal models, a significant hurdle remains largely unaddressed: the challenge of optimally representing disparate data types within a cohesive latent space. This representation is essential for facilitating effective multimodal reasoning, as it allows for the comprehensive analysis and interpretation of data from various sources. The difficulty lies in achieving a representation that not only accommodates the diversity of data types but also preserves the intrinsic relationships among them. This challenge arises because reducing dimensionality results in a lossy compression of our data, while increasing dimensionality leads to a sparse representation of our original data. Finding this balance is of keen interest to our research, as it enables a more nuanced and insightful analysis as well as reduced ``hallucination" type behaviors of modern machine learning approaches.

Therefore in this research, we introduces AlignXpert, an optimization algorithm that draws inspiration from Kernel Canonical Correlation Analysis (CCA). AlignXpert is designed to tackle the intricacies of multimodal data integration by bounding the dimensionality search problem and trying to achieve maximal retention of similarity between the data types, thereby enhancing the quality of data representation. We showcase our optimization problems solution by how are new data representation perform across a set of reasoning tasks. 

\section{Related Work}
\subsection{Data Representation}
Data representation and representation learning are intertwined concepts that have been extensively studied (\citep{bengio2013representation}; \citep{zhong2016overview}; \citep{orejas1981representation}; \citep{neri1994knowledge}) and form the foundation of modern machine learning and artificial intelligence. Data representation involves converting raw data into a structured format that machines can interpret, facilitating effective processing, analysis, and storage of information. This is often achieved by representing the raw data as high-dimensional feature vectors, known as embeddings, which capture the information as accurately as possible. 

Representation learning delves into the process of learning or deriving these structured representations directly from the data itself, using sophisticated algorithms and neural network architectures. This approach moves away from traditional handcrafted feature engineering, aiming instead to automatically discover the most relevant and discriminative features from the raw data. This enhances the performance and generalization capabilities of machine learning models. Through the use of techniques such as autoencoders, convolutional neural networks, and recurrent neural networks, representation learning has revolutionized domains like computer vision, natural language processing, and speech recognition, leading to unprecedented advancements in artificial intelligence (\citep{arora2017provable}).

\subsubsection{Text}

Text representations are a fundamental concept in natural language processing (NLP) and modern machine learning as witnessed from recent popularized Large Language Model research, enabling computers to understand and manipulate human language by representing words, phrases, or entire documents as vectors of numbers. The exploration of NLP technologies commenced with the development of bag-of-words models (\citep{harris1954distributional}; \citep{zhang2010understanding}) primarily focused on the frequency of word occurrence. This foundational approach was subsequently augmented by the adoption of the Term Frequency-Inverse Document Frequency (TF-IDF) technique \citep{sparck2004statistical}, which was a statistical measure designed to assess the significance of a word within a document, relative to a collection or corpus. By taking into account not only the frequency of words but also their distribution across documents, TF-IDF offered a solution to some of the limitations inherent in bag-of-words models. The advent of word embeddings, particularly through methodologies such as Word2Vec (\citep{mikolov2013distributed}; \citep{mikolov2013efficient}; \citep{le2014distributed}) and GloVe \citep{pennington2014glove}, constituted a pivotal advancement in the field. These techniques facilitated the representation of semantic relationships within a dense vector space, thereby capturing the essence of word associations more effectively. 

2017 began a new era of language modelings as transformers \citep{vaswani2017attention} and contextual embeddings, exemplified by models such as ELMo \citep{sarzynska2021detecting} and BERT \citep{devlin2018bert} changed what we know about modern NLP techniques. These models represented a significant leap forward by producing unique word representations that were sensitive to the surrounding context, markedly enhancing the understanding of language nuances. The progression towards large-scale models, including GPT-3 \citep{brown2020language}, T5 \citep{raffel2020exploring}, and XLNet \citep{yang2019xlnet}, marked yet another advance in the field, bringing about improvements in text generation and comprehension capabilities. Now language models are ubiquitous and have been applied on various domains across a range of tasks \citep{bahdanau2014neural, lewis2019bart, wang2023mathcoder, trinh2024solving, lee2024multimodal, radford2019language, lee2024can, hegselmann2023tabllm, dinh2022lift, ono2024text, bahdanau2014neural}

\subsubsection{Vision (Images \& Video)}

The history of vision representations spans from hand-crafted features to deep learning models. Initially, visual recognition tasks relied on manually designed features, such as edges and textures, to represent images and videos. The introduction of SIFT \citep{lowe2004distinctive} and HOG \citep{dalal2005histograms} in the early 2000s marked significant advancements in feature extraction. The breakthrough came with deep learning and convolutional neural networks (CNNs) \citep{lecun1998gradient}, particularly after the success of AlexNet in 2012 \citep{krizhevsky2012imagenet}. CNNs automatically learn hierarchical feature representations from data, significantly improving accuracy in image and video recognition tasks. Recently, the adoption of transformers, originally designed for natural language processing, has made a significant impact on vision tasks, proving to be groundbreaking. Vision transformers (ViT) \citep{dosovitskiy2020image} and ViViT \citep{arnab2021vivit}, aimed specifically at video, showcase cutting-edge methodologies. They utilize self-attention mechanisms to understand global dependencies within images and videos, heralding a major shift in the processing and comprehension of visual data.

\subsubsection{Other Modalities}
Beyond text and vision, representation learning has made significant strides in audio, time series data, and other modalities. In audio, deep learning models like WaveNet \citep{oord2016wavenet} and YAMNet (\textit{Google}) have revolutionized how machines understand and generate human-like speech and music by learning complex patterns within sound waves. For time series, algorithms like Long Short-Term Memory (LSTM) networks \citep{hochreiter1997long} and Temporal Convolutional Networks (TCN) \citep{bai2018empirical} have been pivotal in forecasting and detecting patterns over time, applicable in financial markets, weather prediction, and health monitoring. And recently Mamba \citep{gu2023mamba} was designed to address some limitations of transformer models, particularly in processing long sequences, by incorporating the Structured State Space sequence model. MAMBA introduces enhancements for handling long data sequences efficiently, making it significantly more effective for tasks involving long sequences compared to previous methods. These advancements underline the versatility of representation learning across different data modalities, expanding the frontier of machine understanding.

\subsection{Multimodality}

Multimodality is a more recent area of research presenting a natural view of teaching machines how to utilize various forms of data for reasoning and inference \citep{lahat2015multimodal}. This field acknowledges that the world is rich with diverse information sources, including text, images, audio, and video, and aims to develop models capable of understanding and integrating these different modalities together. The goal of these models is to capture richer representations of the underlying data, leading to more robust and nuanced reasoning capabilities. This approach has gained lots of traction in recent months, where real-world understanding often requires synthesizing information from multiple sources. Some notable frameworks and models in this space include Dall$\cdot$E \citep{ramesh2021zero}, Google Gemini \citep{team2023gemini}, and much much more.

\subsubsection{Alignment}
One of the most popular forms of unifying multimodal data is through modality alignment. Alignment, involves projecting or transforming the embeddings from one modality's latent space into another's, or into a shared latent space, so that the embeddings from different modalities can be directly compared or combined. These mechanisms could be as simple as a Linear Projection, or Cross modal attention. Many creative ways have been implemented on this front utilizing self attention layers \citep{lee2024multimodal}, projections \citep{radford2021learning} and more \citep{yun2022modality}. More interestingly \citep{zhou2024aligning} highlights how a lack of good alignment can often lead to hallucinations in multimodality required a need for better multimodal data representation.

\subsubsection{Contrastive Learning}

Contrastive learning is a transformative approach in machine learning that leverages the concept of contrasting positive examples against negative ones to learn useful data representations without extensive labeled datasets. Similarly, in multimodal datasets, they have found ways to use this appraoch to learn positive and negative pairs from different modalities (e.g. text-to-image, image-to-sound, etc.)

Among the standout methods in this field, CLIP (Contrastive Language–Image Pre-training) by OpenAI \citep{radford2021learning} is renowned for its ability to learn visual concepts from natural language descriptions, thereby bridging the gap between images and texts with remarkable flexibility. Similarly, ImageBind \citep{girdhar2023imagebind} and other models like SimCLR \citep{chen2020simple} and MoCo \citep{he2020momentum} further enrich the landscape of contrastive learning by enhancing the efficiency and effectiveness of unsupervised learning in visual representation and beyond. 

\section{Methodology}

\begin{figure}[ht]
    \centering
    \includegraphics[width=0.7\columnwidth]{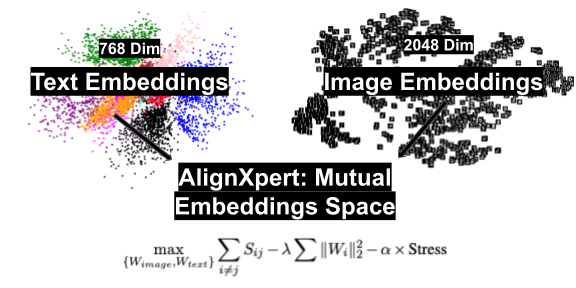}
    \caption{An overview of our proposed methodology which shows two modalities in this case images and text being projected into the optimal dimensionality}
    \label{method}
\end{figure}

The AlignXpert algorithm is introduced as an optimization problem aimed at finding the optimal representation between two different modalities. This is achieved by identifying the highest similarity when projected into bounded dimensionalities between the two modality vectors, while imposing additional constraints. Traditionally, researchers have projected one modality into another's latent space for alignment. However, we question how the introduction of sparsity or the reduction of a data's original representation affects the overall performance of modern machine learning models. We hypothesize that a combination of these two operations may be necessary to find an optimal alignment.

Therefore, in the subsequent subsections, we delve into the conceptual framework and practical application of AlignXpert. We will explore the pedagogical principles that inform the design of AlignXpert, illustrating how these principles underpin the method's approach. Furthermore, we will outline our evaluation framework to assess the effectiveness of our optimization method.

\subsection{Kernel Canonical Correlation Analysis (Kernel CCA)}

Kernel Canonical Correlation Analysis (Kernel CCA) \citep{gonen2011multiple} enhances traditional Canonical Correlation Analysis (CCA) \citep{hotelling1992relations} by using kernel methods to uncover nonlinear relationships between multidimensional variable sets. Unlike CCA that searches for linear correlations in the original data, Kernel CCA employs a kernel function to map data into a higher-dimensional feature space, revealing nonlinear patterns. The method focuses on maximizing the correlation between linear combinations of the mapped data (\(U = a^T\Phi(X)\) and \(V = b^T\Phi(Y)\)) through kernel functions \(K_X\) and \(K_Y\), aiming to optimize the correlation:
\[
\rho = \frac{a^T\Phi(X)^T\Phi(Y)b}{\sqrt{a^T\Phi(X)^T\Phi(X)a} \sqrt{b^T\Phi(Y)^T\Phi(Y)b}}
\]
This approach allows for the exploration of complex, nonlinear associations, expanding CCA's utility beyond linear contexts.

\subsection*{AlignXpert Overview}

AlignXpert leverages the principles of Kernel Canonical Correlation Analysis (Kernel CCA) while introducing a nuanced approach to multimodal data analysis. This novel variant, informed by an advanced optimization function, targets the intricate dynamics of datasets across diverse types—such as images, text, and audio—channeling them into a cohesive, shared subspace. The revised strategy transitions from a correlation-centric analysis to a comprehensive evaluation of similarity, regularization, and stress minimization. Here, we outline the AlignXpert methodology, emphasizing its refined focus on optimizing similarity alongside dimensionality reduction.

\subsection*{Kernel Transformation and Multi-Modal Mapping}

At the heart of AlignXpert is the strategic use of kernel functions \(K_i\) to transform data from its native embedding space into an expanded, higher-dimensional feature space. Each modality \(X_i\) undergoes a kernel-induced mapping \(\Phi_i: X_i \rightarrow \mathcal{H}_i\), with \(\mathcal{H}_i\) denoting a Hilbert space. This transformation reveals nonlinear associations and patterns previously obscured within the original data dimensions, as observed in many high-fidelity embeddings.

\subsection*{Objective Function Formulation}

The foundation of AlignXpert is a dual-objective function that aims to maximize similarity across modalities within the feature space while minimizing the dimensionality of the shared subspace. Unlike traditional Kernel CCA, which focuses on maximizing pairwise correlations (\(\rho_{ij}\)), AlignXpert calculates the similarity between embeddings, refining the analytical lens through which multimodal interactions are assessed:

\[
\text{Similarity} = \sum_{i \neq j} S_{ij} = \sum_{i \neq j} \text{kernel}(\Phi_i(X_i), \Phi_j(X_j))
\]

Here, \(S_{ij}\) represents the similarity measure between the transformed embeddings of modalities \(X_i\) and \(X_j\), facilitated by a specified kernel function. AlignXpert incorporates a regularization component to control the solution's complexity, aiming for a streamlined, lower-dimensional subspace. This regularization, combined with an innovative stress evaluation metric, ensures the alignment’s fidelity to the original data's spatial structure, enhancing both analytical depth and computational efficiency.

\subsection*{Optimization}

The optimization framework of AlignXpert carefully balances enhancing inter-modal similarity with the necessity of dimensionality restraint. The revised optimization objective is articulated as follows:

\[
\max_{\{W_{image}, W_{text}\}} \sum_{i \neq j} S_{ij} - \lambda \sum \|W_i\|_2^2 - \alpha \times \text{Stress}
\]

In this formulation, \(W_i\) represents the weights of the projection layers for each modality, \(\lambda\) is the regularization parameter, and \(\alpha\) indicates the weight given to the stress component. The stress term quantifies the discrepancy between the original and projected distances, ensuring the preservation of intrinsic data geometries within the reduced subspace.

\subsection*{Stress Metric}

We introduce a stress metric aimed at calculating the distances between matrices \(D\) and \(D'\), representing the original and projected embeddings, respectively. This metric quantifies the alterations in geometric relationships induced by the projection, providing insight into the impact of dimensionality reduction.

\[ \text{Stress} = \sqrt{\frac{\sum (d_{ij} - d'_{ij})^2}{\sum d_{ij}^2}} \]

where \(d_{ij}\) and \(d'_{ij}\) are the distances between elements \(i\) and \(j\) in the original and projected spaces, respectively.

\subsection*{AlignXpert Algorithm}

In our algorithm, we wish to highlight the optional parameters that can be tuned according to the desired emphasis: Learning Rate, num\_epochs, Dimension Range, Regularization Parameter, and Stress Weight. We provide detailed information on these parameters in our appendices.

\begin{algorithm}
\caption{AlignXpert Algorithm}
\begin{algorithmic}[1]
\Procedure{Optimize Alignment}{}
\While{not converged}
\State Project embeddings: $\Phi_{image}, \Phi_{text}$
\For{each pair $(i, j), i \neq j$}
\State Compute similarity $S_{ij}$ using a kernel function
\EndFor
\State Compute mean similarity $\bar{S}$
\State Compute distances in original and projected spaces $D, D'$
\State Compute stress $\text{Stress}$
\State Compute regularization $R$
\State $L \gets -\bar{S} + R + \alpha \times \text{Stress}$
\State Update $\{W_{image}, W_{text}\}$ to minimize $L$
\EndWhile
\State \Return Optimal $\{W_{image}, W_{text}\}$
\EndProcedure
\end{algorithmic}
\end{algorithm}

\subsection{Evaluation Metrics}

We evaluate the performance of AlignExpert by conducting a thorough analysis at the embedding level, followed by its application in two downstream tasks: retrieval and classification. \\

\noindent\textbf{Embedding Analysis}:\\
Our evaluation begins by analyzing the effects of lossy compression and introduced sparsity on each modality, essential for gauging how much data can be compressed without significant information loss. This step parallels dimensionality reduction methods like PCA, which use explained variance ratios to evaluate information loss.

We then apply the AlignExpert algorithm to determine optimization outcomes and modality similarities. Next, we assess the optimal embeddings' performance in various alignment scenarios, including retrieval and classification tasks, to measure their real-world applicability. To benchmark AlignExpert's effectiveness, we compare it with three models using different alignment strategies: one employs dimensionality reduction and aligns in the text space for efficient representation assessment; another uses sparsity to project lower-dimensional modalities into higher-dimensional spaces; and the third, unique to AlignExpert, evaluates performance metrics within the bounds of the lowest and highest dimensional modalities. \\ 

\noindent\textbf{Retrieval}\\
In our retrieval task, we employ a small Pokémon Dataset which contain images and a visual description to simulate an unsupervised retrieval problem, aiming to assess the quality of the output. By inputting content from both modalities, we evaluate the relevance of the returns using Precision@k, Recall@k, and f1 metrics. These metrics are based on queries with trivial solutions, such as colors and visual characteristic features. For this analysis, we leverage a pretrained CLIP model as the backbone framework using its encoder models for both text and images for our retrieval, enabling us to harness its powerful feature extraction capabilities.\\

\noindent\textbf{Classification}\\
In our classification component, we address the evaluation of our model's performance in multi-class classification tasks through established metrics. We developed a multimodal sentiment analysis framework that integrates Google's Vision Transformer \citep{wu2020visual}, \citep{deng2009imagenet} for image processing and the sentence transformer model (\texttt{sentence-transformers/all-mpnet-base-v2\\-ep10}) for text encoding. These pretrained models are utilized to extract pertinent features from both images and text, which are subsequently merged and processed through a series of fully connected layers for the purpose of sentiment classification. The training and evaluation of our model were facilitated using the Weights \& Biases providing comprehensive tracking and optimization of our experiments. To thoroughly understand our model's capability, we employed a range of metrics, including micro and macro F1 scores, AUROC (Area Under the Receiver Operating Characteristic curve), and Accuracy and conduct thorough analysis as indicated by \cite{lee2024feetframeworkevaluatingembedding}. 

\section{Results}

\subsection{Measuring the Lossy compression of Projections}

\begin{figure}[ht]
    \centering
    \includegraphics[width=0.7\columnwidth]{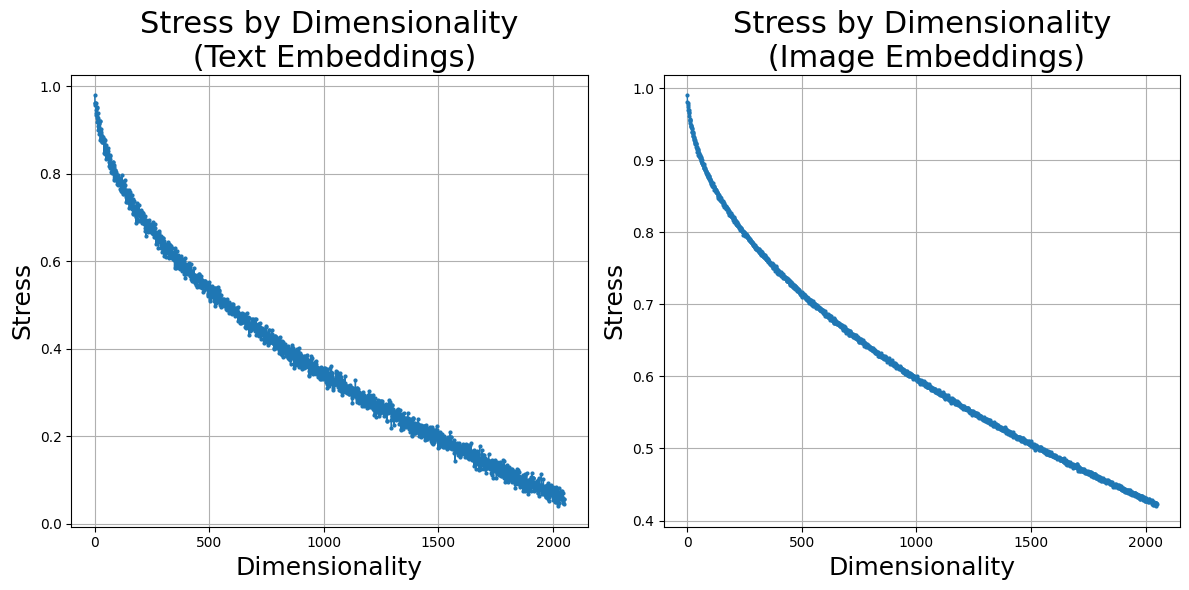}
    \caption{The Stress of Word \& Image Embeddings as a function of dimensionality. \textit{\textbf{Note:} Y-axis are different scales.}}
    \label{fig1}
\end{figure}

Our investigation commenced with an examination of information loss in word embeddings during dimensionality transitions, focusing on fidelity and distortion through similarity metrics. Due to the inability to compute inner products between vectors of differing dimensions, we devised a stress metric for this purpose. Employing a linear projection method, our findings, as depicted in Figure \ref{fig1}, revealed significant information loss with dimensionality reduction. Intriguingly, projecting text embeddings into dimensions higher than their original ones still resulted in a decreasing stress value.

\subsection{AlignXpert}

\subsubsection*{Case Study: Retrieval}

\begin{figure}[ht]
    \centering
    \includegraphics[width=0.7\columnwidth]{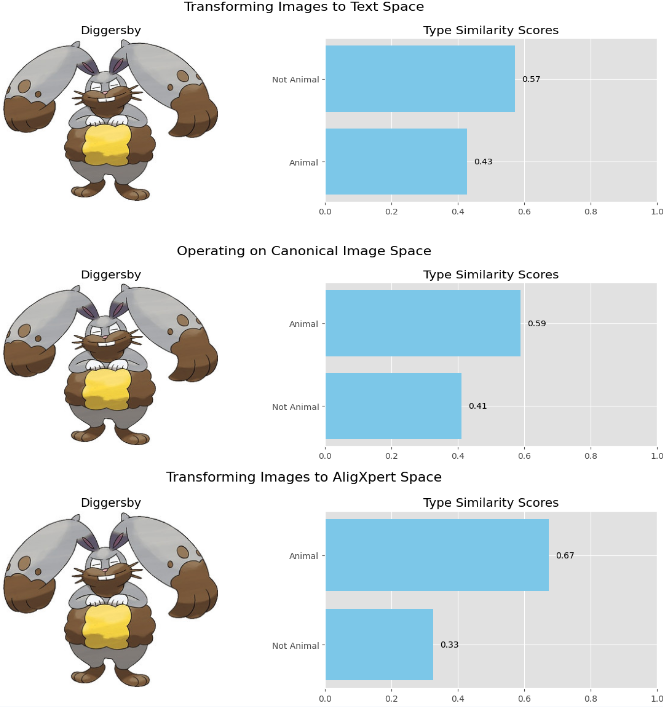}
    \caption{\textbf{Text Retrieval}: In this figure we present a Pokemon based on what appears to be a bunny. We see how varying projection strategies on our Images and Text affect retrieval Performance. We also see AlignXpert appears to maximize its confidence that it is an Animal Based Pokemon}
    \label{poke1}
\end{figure}

In a practical demonstration involving Pokémon data, we illustrated AlignXpert's utility in retrieval tasks. Figure \ref{poke1} showcases the model's efficacy in discerning whether a depicted Pokémon is animal-inspired based on image input (we discern animal inspired design based on an online Pokemon encylopedia). This task highlighted the challenges of projecting from an image space of 2048 dimensions to a text space of 768 dimensions, where information loss sometimes led to inaccuracies or ``hallucinations''. We ran 5 independent text retrieval tasks to obtain our results.

\begin{figure}[ht]
    \centering
    \includegraphics[width=0.5\columnwidth]{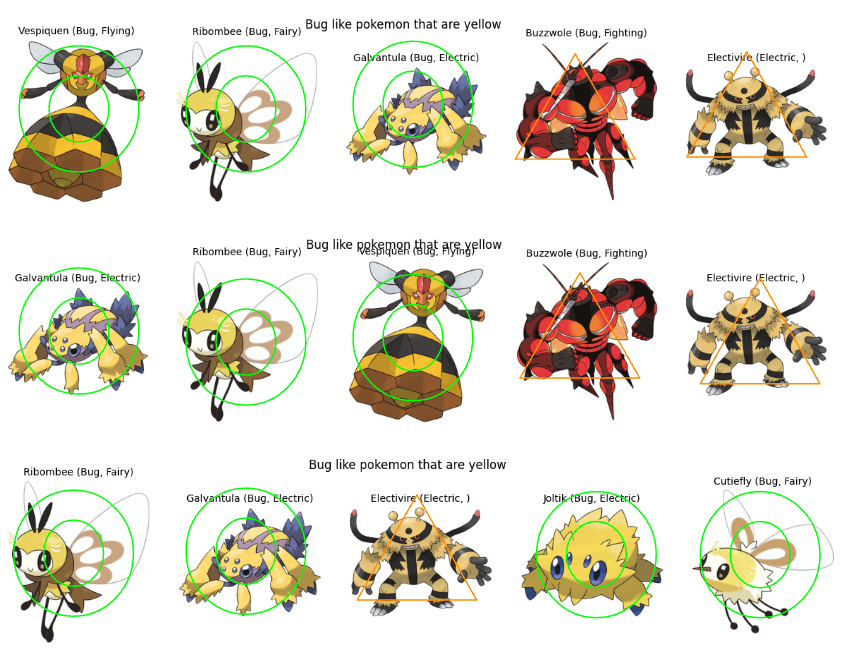}
    \caption{\textbf{Image Retrieval}: In this figure we present a textual prompt and want to receive all relevant images related to the prompt. The prompt reads: "\textit{Bug like pokemon that are yellow}". We also add green circles for good matches, orange triangles for semi-matches meeting one of the conditions. Similar to Figure 1, from top to bottom we have text in its canonical form, text projected into image space, and text projected into AlignXpert Space. We see that AlignXpert is able to retrieve the most green circles in this example.}
    \label{poke2}
\end{figure}

\begin{figure}[ht]
    \centering
    \includegraphics[width=0.5\columnwidth]{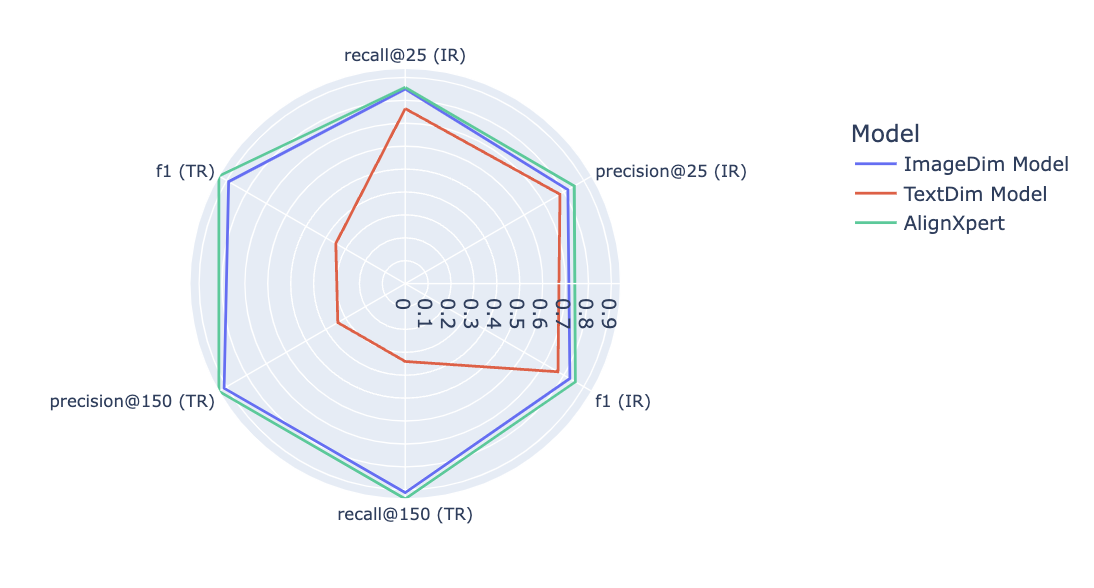}
    \caption{A spider plot scoring the Recall@k, Precision@k, and F1 of both Text Retrieval (TR) and Image Retrieval Tasks. Scores seen are averages against multiple queries. We see negative effects when images are projected down but minimal effects when text is projected up as seen in the ImageDim Model.}
    \label{poke11}
\end{figure}

Conversely, an image retrieval task prompted by text (Figure \ref{poke2}) demonstrated minimal performance disparity between text projected into image space and the use of a canonical image space model. AlignXpert exhibited marginally superior performance in this regard. To further rigorously evaluate these models on these unsupervised tasks, we assessed their ability to retrieve relevant matches through 10 independent experiments in image retrieval, in addition to the 5 independent experiments in text retrieval. These results, as seen in Figure \ref{poke11}, reflect our findings.

\subsubsection*{Case Study: Classification}

\begin{figure}[ht]
    \centering
    \includegraphics[width=0.5\columnwidth]{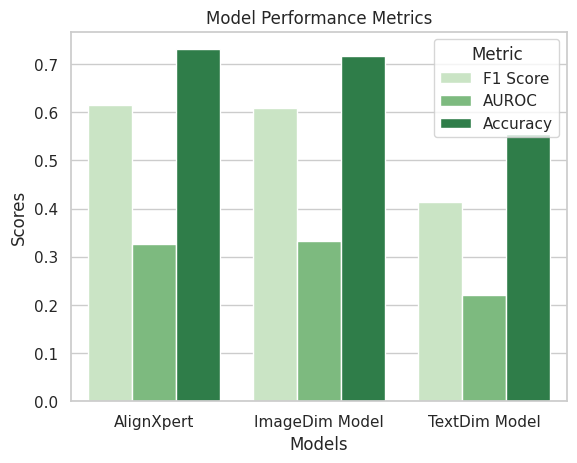}
    \caption{A bar graph showcases the benchmarks of the classification task. AlignXpert opted for a dimensionality of 2038 to achieve optimal alignment. Overall, the performance gains are very marginal when compared to the Image Embedding Model, yet significant in contrast to the Text Embedding Model.}
    \label{bar}
\end{figure}

A more robust demonstration of these projections is provided by evaluating them on a classification task, which offers more stable performance metrics for assessing AlignXpert's effectiveness. We tasked each model with predicting binary classifications in multimodal sentiment analysis. The dataset employed focuses on internet memes for emotion analysis, specifically targeting sentiment. It comprises 8,000 memes, each annotated with human-assessed sentiment tags. Consequently, we trained a multimodal model using various projection strategies to predict whether each sample is "offensive" or "not offensive." Our results are evaluated and presented in Figure \ref{bar}.

\noindent Generally, we observe that AlignXpert yields marginal improvements in both classification and retrieval tasks.

\section{Discussion}

\subsection{AlignXpert Prefers Projecting Upwards}
Often, when we operated AlignXpert with a minimal focus on dimensionality reduction, we observed that lower-dimensional embeddings were typically projected upwards, aligning closely with the space of higher-dimensional embeddings. This tendency correlates with the performance issues we encountered; reducing the dimensionality of higher-dimensional embeddings led to a loss of similarity between the multimodal data. Consequently, users of AlignXpert may prioritize dimensionality reduction in their optimization problem but should be mindful of this characteristic tendency.

\subsection{AlignXpert's Trade-off}

Our research identified a trade-off between performance and dimensionality reduction in optimization. Multimodal analysis demonstrated that upward projection of a modality incurs minimal information loss, as shown in Figure \ref{fig1}. Nevertheless, this approach increases sparsity and computational complexity due to added dimensions. Upon reflection, while dimension reduction may compromise performance, it invariably results in some information loss from the original representation. Hence, in scenarios involving two modalities, the modality with smaller dimensionality is often projected upwards to enhance similarity between the modalities.

\subsection{Emerging Properties of AlignXpert}

\subsubsection*{Modality-Agnostic Capability}
A notable attribute of AlignXpert is its modality-agnostic capability, effectively demonstrated in various dataset pairings, including text-to-image and others. Crucially, AlignXpert's adaptability spans a wide range of modalities, provided they can be represented in vector form. This versatility ensures its applicability across a broad spectrum of multimodal research fields.

\section{Conclusion}

In this paper, we introduced AlignXpert, an optimization algorithm designed to synergize alignment strategies by maximizing similarities while incorporating dimensionality reduction and geometric constraints. This innovation marks a significant contribution to the field, especially with its novel integration of dimensionality reduction and geometric constraints into multimodal alignment strategies.

We demonstrated that AlignXpert facilitates performance enhancements in both retrieval and classification tasks. This result is particularly significant, suggesting that optimal alignment involves a combination of dimensionality reduction and upward projection to achieve maximal performance. One of our main findings from this study was that AlignXpert, when set to default parameters, tended to project upwards towards the image modality in both analyses. This observation indicates that projecting into a reduced space has a more detrimental effect on maintaining similarity between modalities. We posit that AlignXpert represents a promising avenue for enhancing the efficacy of multimodal research, with potential applications spanning a wide range of fields, from healthcare to finance and more. We encourage researchers to rigorously test our work, as different datasets from different domains present new challenges in multimodal analysis.

\subsection{Future Work}

Future directions for this work involve rigorously testing our method across multimodal datasets from diverse domains. This underscores the crucial need for the distribution of multimodal datasets that can serve as standard benchmarks in the field, similar to how the MNIST or IRIS datasets have been used in early machine learning research. Although we have utilized publicly available datasets, we are keen to explore proprietary datasets across various domains. It is also vital to apply our method to a broader array of data modalities, extending beyond merely text and images, and to conduct more comprehensive testing. Furthermore, investigating unconventional modality combinations, especially those prevalent in medical and scientific research, offers intriguing possibilities for future exploration. Such an approach not only expands the applicability of our work but also contributes significantly to the advancement of multimodal research.

\subsubsection*{Beyond Bivariate Cases}
A larger extension of this work involves broadening our examples to include three or more modalities. As multimodal research advances, datasets extending beyond bivariate configurations will become increasingly common. Thus, it is essential to adapt and enhance AlignXpert in future iterations to handle interactions among multiple modalities, moving beyond merely binary modality scenarios. We present the mathematical foundation for this extension here:

\subsubsection*{Generalization to \(n\) Modalities}
In scenarios involving \(n\) modalities, AlignXpert's optimization framework is designed to encompass an arbitrary number of modalities. Its aim is to maximize inter-modal similarity, enforce regularization, and minimize stress across the dataset:

\[
\max_{\{W_i\}} \sum_{\substack{i, j=1 \\ i \neq j}}^n S_{ij} - \lambda \sum_{i=1}^n \|W_i\|_2^2 - \alpha \times \text{Stress}
\]

This objective function seeks to identify a shared lower-dimensional space that enhances the weighted similarities (\(S_{ij}\)) across all modality pairs, incorporates regularization to promote compactness of the representations (\(\|W_i\|_2^2\)), and minimizes stress to preserve the integrity of original distances in the reduced space. This advancement signifies a notable progression from its initial version, allowing for more detailed analyses of intricate multimodal datasets.

\bibliography{colm2025_conference}
\bibliographystyle{colm2025_conference}

\appendix
\onecolumn
\section*{Appendix}
\label{sec:appendix}

Sections A-C provide the necessary mathematical background required to understand some of the concepts presented in this paper. They highlight key definitions and proofs essential for grasping these concepts. Section C includes the hyperparameters of the AlignXperts model. Section D showcases additional results from the paper, helping to contextualize our findings.

\section{Hilbert Spaces \& Latent Spaces}

Hilbert spaces and latent spaces are foundational concepts in mathematics and machine learning. This section \ref{sec:appendix} explores the mathematical structures of Hilbert spaces, and the role of latent spaces in machine learning.

\section*{Hilbert Spaces: A Mathematical Framework}

A \textit{Hilbert space} is a complete inner product space that extends the methods of vector algebra and calculus to spaces with any finite or infinite number of dimensions. A Hilbert space $\mathcal{H}$ is characterized by the following properties:

\begin{enumerate}
    \item \textbf{Vector Space:} It is a vector space over the field of real or complex numbers, equipped with vector addition and scalar multiplication.
    \item \textbf{Inner Product:} There exists an inner product $\langle x, y \rangle$ for vectors $x, y$ in $\mathcal{H}$, satisfying:
    \begin{itemize}
        \item Conjugate symmetry: $\langle x, y \rangle = \overline{\langle y, x \rangle}$
        \item Linearity in the first argument: $\langle ax + by, z \rangle = a\langle x, z \rangle + b\langle y, z \rangle$
        \item Positive-definiteness: $\langle x, x \rangle \geq 0$ with equality iff $x = 0$
    \end{itemize}
    \item \textbf{Completeness:} Every Cauchy sequence in the space converges to a limit within the space.
\end{enumerate}

The norm $\|x\|$ of a vector $x$ in $\mathcal{H}$, derived from the inner product, is defined as $\sqrt{\langle x, x \rangle}$. This norm induces a metric and defines the topology of the space.

\section*{Latent Spaces in Machine Learning}

Latent spaces in machine learning are abstract, high-dimensional spaces designed to capture intrinsic patterns and features of data. These spaces leverage the properties of Hilbert spaces for efficient data encoding and manipulation.

\subsection*{Dimensionality and Encoding}

Machine learning models, such as autoencoders, encode input data into a lower-dimensional latent space to preserve relevant information. This process mirrors projecting vectors onto a subspace in a Hilbert space, maintaining the geometric relationships of the data.

\subsection*{Optimization and Function Spaces}

Optimization in machine learning often occurs in function spaces, which are Hilbert spaces. Kernel methods, exploiting the Hilbert space structure, ensure linear separability in a high-dimensional space. The Representer Theorem guarantees the existence of an optimal solution in the space spanned by data points.

\subsection*{Mathematical Representation of Latent Spaces}

Let $\mathcal{L}$ represent a latent space and $\mathcal{D}$ a high-dimensional data space. A mapping $f: \mathcal{L} \rightarrow \mathcal{D}$ by a machine learning model allows for the representation of data points or features in $\mathcal{D}$ for any latent vector $z \in \mathcal{L}$. The Hilbert space structure facilitates efficient gradient computation and optimization of loss functions.

\section{Hilbert Spaces and RKHS}

\begin{definition}[Hilbert Space]
We defined a \emph{Hilbert Space} $\mathcal{H}$ to be a complete inner product space equipped with an inner product $\langle \cdot, \cdot \rangle$, inducing a norm $\|f\| = \sqrt{\langle f, f \rangle}$ for all $f \in \mathcal{H}$.
\end{definition}

\begin{definition}[Reproducing Kernel Hilbert Space]
A Reproducing Kernel Hilbert Space \emph{(RKHS)} over a set $X$ is a Hilbert space $\mathcal{H}$ of functions $f: X \rightarrow \mathbb{R}$ such that for every $x \in X$, the evaluation functional $L_x(f) = f(x)$ is continuous. There exists a function $k: X \times X \rightarrow \mathbb{R}$, known as the reproducing kernel, satisfying $k(x, \cdot) \in \mathcal{H}$ and $f(x) = \langle f, k(x, \cdot) \rangle_{\mathcal{H}}$ for all $f \in \mathcal{H}$ and $x \in X$.
\end{definition}

In other words a RKHS is essentially a special kind of mathematical space filled with functions that map elements from a set \(X\) to real numbers (\(\mathbb{R}\)). This space has two key features. First, you can pick any point \(x\) from \(X\) and evaluate any function \(f\) from this space at \(x\) smoothly—that is, without any abrupt changes (this is what's meant by the evaluation functional \(L_x(f) = f(x)\) being continuous). Second, there's a special function \(k\), called the reproducing kernel, that operates on pairs of points from \(X\). This kernel function has a neat property: for any point \(x\) in \(X\) and any function \(f\) from the space, evaluating \(f\) at \(x\) is the same as doing an inner product between \(f\) and the kernel function \(k\) with one of its inputs fixed at \(x\). This reproducing kernel captures the essence of the space, allowing every function within the space to be "reproduced" using just this kernel function and the inner product operation.

\subsection*{Regularized Empirical Risk Minimization}

Given a dataset $\{(x_1, y_1), \ldots, (x_n, y_n)\} \subseteq X \times \mathbb{R}$, the regularized empirical risk minimization problem seeks a function $f \in \mathcal{H}$ that minimizes
\[
J(f) = \sum_{i=1}^{n} L(y_i, f(x_i)) + \lambda \|f\|^2_{\mathcal{H}},
\]
where $L: \mathbb{R} \times \mathbb{R} \rightarrow [0, \infty)$ is a loss function and $\lambda > 0$ is a regularization parameter. We see that this minimization function looks very familiar to our optimization function in the main paper.

\section{The Representer Theorem: Simplifying Machine Learning Solutions}

The Representer Theorem stands as a cornerstone in the realms of machine learning and statistical learning theory, illustrating the profound efficacy of kernel methods. At its core, the theorem reveals that solutions to a vast spectrum of optimization challenges, particularly those involving the sophisticated structure of Reproducing Kernel Hilbert Spaces (RKHS), can be distilled into simpler expressions. More precisely, these solutions manifest as weighted sums of kernel functions—measures of similarity—evaluated over the training data points. This simplification not only renders formidable problems more computationally tractable but also deepens our theoretical grasp of why kernel methods are remarkably potent.

Kernel methods serve as a transformative tool in machine learning, adeptly mapping data into an expanded, sometimes boundlessly dimensional, space without necessitating direct computation of these new dimensions. This ingenious approach enables the application of linear solution strategies to inherently nonlinear dilemmas by operating within this "latent" space—a realm where data relationships are more discernibly structured. The Representer Theorem validates this strategy, showing that solutions remain intimately connected to and expressible through the training dataset, even within the complexity of high-dimensional spaces. This theorem is pivotal, affirming that the intricate spaces and mappings employed in machine learning extend beyond theoretical elegance to offer tangible benefits for solving practical problems.

\subsection*{Delving into the Problem}

Imagine a dataset comprising pairs \((x_1, y_1), \ldots, (x_n, y_n)\), where each \(x_i\) belongs to a set \(\mathcal{X}\) and each \(y_i\) to the real numbers \(\mathbb{R}\). Within this setting, \(\mathcal{H}\) represents a RKHS of functions mapping \(\mathcal{X}\) to \(\mathbb{R}\), associated with a kernel \(K\). The objective? To minimize the regularized empirical risk:

\[
\min_{f \in \mathcal{H}} \left\{ \sum_{i=1}^{n} L(y_i, f(x_i)) + \lambda \|f\|^2_{\mathcal{H}} \right\},
\]

guided by a loss function \(L\) and regularization parameter \(\lambda\).

\subsection*{The Essence of the Representer Theorem}

\begin{theorem}[Representer Theorem]
Given a set \(X\), an RKHS \(\mathcal{H}\) over \(X\) with kernel \(k\), a dataset \(\{(x_1, y_1), \ldots, (x_n, y_n)\}\), and a regularization parameter \(\lambda > 0\), any optimal solution \(f^*\) minimizes the functional \(J(f)\) and can be expressed as:

\[
f^*(x) = \sum_{i=1}^{n} \alpha_i k(x_i, x),
\]

where each \(\alpha_i\) is a real number for \(i=1, \ldots, n\).
\end{theorem}

\subsection*{Proof of the Representer Theorem}

The foundation of the proof rests on the unique characteristics of Reproducing Kernel Hilbert Spaces (RKHS), notably the continuity of evaluation functionals and the application of the Riesz Representation Theorem. We approach any function \(f\) within \(\mathcal{H}\) by segregating it into two components: one is a linear amalgamation of kernel functions centered around the training data points, and the other is orthogonal to this collective span. The orthogonality principle, combined with the optimization problem's framework, directs us to the insight that the optimal solution, or minimizer, must be encapsulated within the kernel functions' span focused on the training data, which substantiates the theorem's claim.

The optimal solution \(f^*\) for the given optimization challenge is expressible in the format:

\[
f^*(x) = \sum_{i=1}^{n} \alpha_i K(x_i, x)
\]

where each \(\alpha_i\) is a real number for \(i=1, \ldots, n\).

\subsection*{Proof}

\begin{enumerate}
    \item \textbf{RKHS Characteristics:} Within an RKHS, any function \(f\) is such that its evaluation at any point \(x\), denoted \(L_x(f) = f(x)\), is continuous. The Riesz Representation Theorem assures us of a unique \(\phi_x\) within \(\mathcal{H}\) satisfying \(f(x) = \langle f, \phi_x \rangle_{\mathcal{H}}\) for every \(x\) in \(\mathcal{X}\). Here, \(\phi_x\) manifests as the kernel function, specifically, \(\phi_x(y) = K(x, y)\).
    \item \textbf{Dissecting the Function: }Any chosen function \(f\) in \(\mathcal{H}\) can be dissected into \(f = f_{\parallel} + f_{\perp}\), where \(f_{\parallel}\) is the portion lying within the kernel functions' span at the training data points, represented as \(\sum_{i=1}^{n} \alpha_i K(x_i, \cdot)\), and \(f_{\perp}\) is the segment orthogonal to this span.
    \item \textbf{Orthogonality and Optimization:} The regularization term \(\|f\|^2_{\mathcal{H}}\) benefits from orthogonality, allowing its decomposition, with the orthogonal component \(f_{\perp}\) invariably adding a non-negative value. Given that the loss function's calculations are solely reliant on \(f\)'s values at the training points, \(f_{\perp}\) doesn't aid in reducing the loss but contributes to the regularization penalty.
    \item \textbf{Achieving Minimization:} From the above, it's evident that the objective's minimization is attainable sans the orthogonal component \(f_{\perp}\). Hence, the optimal solution \(f^*\) must be fully represented within the span of the kernel functions evaluated at the training points, culminating in the formula \(f^*(x) = \sum_{i=1}^{n} \alpha_i K(x_i, x)\).
\end{enumerate}

\subsection{Hardware}

We run all our analysis on a Tesla T4 GPU and have written everything in the Python programming language making use of PyTorch to write AlignXpert. Additional Optimization parameters will be listed in the appendix.

\section{Hyperparameters of AlignXpert}

\begin{tabular}{ |p{3cm}||p{3cm}|p{3cm}|p{3cm}|  }
 \hline
 \multicolumn{4}{|c|}{AlignExpert Hyperparameters} \\
 \hline
 Hyperparameter & Default Value & Description & Notes \\
 \hline
 Learning Rate & 0.001 & Controls the speed at which the algorithm updates weights during optimization. & - \\
 \hline
 Epochs & 1000 & Number of times the optimization algorithm will work through.  & Shorter iterations may result to less robust solutions \\
 \hline
 Dimension Range & Bounded by Modalitiy 1's dimensionality and Modality 2's dimensionality& Range from the number of dimensions in image embeddings to the sum of dimensions in image and text embeddings. & Adjust based on specific data. \\
 \hline
 Regularization Parameter & 0.001 & A metric to measure the degree of importance to which you want to reduce the dimensions & - \\
 \hline
 Stress Weight & 0.1 & Influences the importance of our stress metric in the optimization process. & - \\
 \hline
\end{tabular}

\newpage

\section{Additional Results Section}

In this section, we highlight further examples of our analysis that did not make it into the main paper.

\subsection{Retrieval Prompts}
In this portion of the text, we highlight more of the experiments conducted to derive some of the performance metrics from our retrieval system. We base most tasks in both retrieval settings on Pokémon features, as a more sophisticated analysis would require a more holistic dataset. We also gather contextual information, such as whether a Pokémon was based on an animal, by scraping information from crowdsourced sources like \href{https://bulbapedia.bulbagarden.net/wiki/User:TwistedMeow/Pok%C3%A9mon_designed_after_animals}{Bulbapedia}. We also consider all Pokémon not designed from animals to be considered \texttt{``mythical''}.  Other tasks are considered trivial based on the design of the Pokémon.

\subsubsection*{Image Retrieval}
\begin{table}[htbp]
\centering
\caption{Queries entered to derive the F1 score, precision@k, and recall@k for the image retrieval tasks where \(k=25\)}
\label{tab:pokemon_queries}
\begin{tabular}{ |p{3cm}||p{3cm}|p{3cm}|p{3cm}|  }
 \hline
 \multicolumn{4}{|c|}{Pokémon Query List} \\
 \hline
 \multirow{10}{*}{Experiment} & \multicolumn{3}{c|}{Query Prompt} \\
 \cline{2-4}
 & \multicolumn{3}{p{9cm}|}{1. \texttt{Pokémon That Look Like Dragons}} \\
 & \multicolumn{3}{p{9cm}|}{2. \texttt{Blue Pokémon}} \\
 & \multicolumn{3}{p{9cm}|}{3. \texttt{Animal Pokémon}} \\
 & \multicolumn{3}{p{9cm}|}{4. \texttt{Pokémon That Look Like Birds}} \\
 & \multicolumn{3}{p{9cm}|}{5. \texttt{Pokémon That Look Like They Live in the Water}} \\
 & \multicolumn{3}{p{9cm}|}{6. \texttt{Pokémon That Are Yellow}} \\
 & \multicolumn{3}{p{9cm}|}{7. \texttt{Pokémon That Look Like Mythical Creatures and Not Like Animals}} \\
 & \multicolumn{3}{p{9cm}|}{8. \texttt{Flying Pokémon}} \\
 & \multicolumn{3}{p{9cm}|}{9. \texttt{Bug-Like Pokémon That Are Yellow}} \\
 & \multicolumn{3}{p{9cm}|}{10. \texttt{Large Pokémon}} \\
 \hline
\end{tabular}
\end{table}

\subsubsection*{Text Retrieval}

\begin{table}[htbp]
\centering
\caption{The example queries for text based retrieval. We based tasks on Pokémon based on specific characteristics. We also increased $k$ to be $150$ to score the metrics on text retrieval tasks. Text Retrieval can also be seen here as binary classification.}
\label{tab:pokemon_characteristics}
\begin{tabular}{|l||l|l|}
 \hline
 \multicolumn{3}{|c|}{Pokémon Query Prompts} \\
 \hline
\textbf{Characteristics} & \textbf{Has the Characteristic} & \textbf{Does Not Have the Characteristic} \\
\hline
\texttt{This Pokémon is based on} & \texttt{An Animal} & \texttt{Not An Animal} \\
\hline
\texttt{This Pokémon} & \texttt{Has Wings} & \texttt{Does Not Have Wings} \\
\hline
\texttt{This Pokémon is} & \texttt{Blue} & \texttt{Not Blue} \\
\hline
\texttt{This Pokémon is} & \texttt{Bird-like} & \texttt{Not Bird-like} \\
\hline
\texttt{This Pokémon has a} & \texttt{Mythical Design} & \texttt{Not Mythical Design} \\
\hline
\end{tabular}
\end{table}


\begin{figure}
    \centering
    \includegraphics[width=\textwidth]{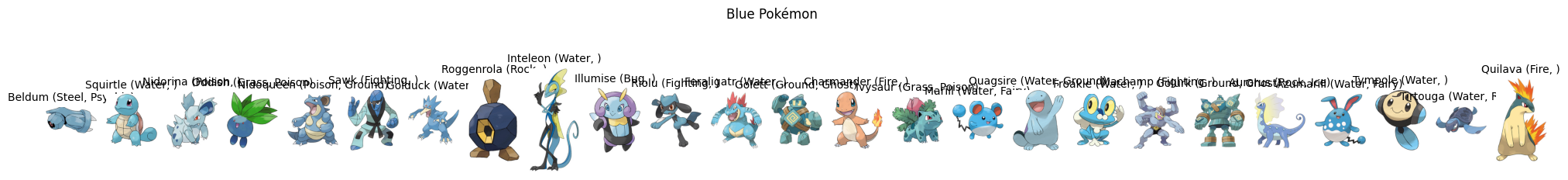}
        \includegraphics[width=\textwidth]{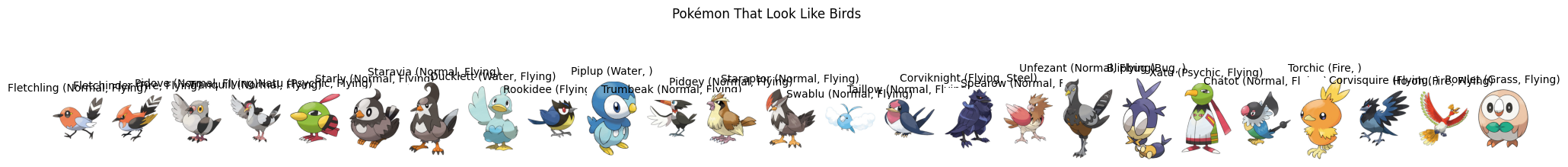}
            \includegraphics[width=\textwidth]{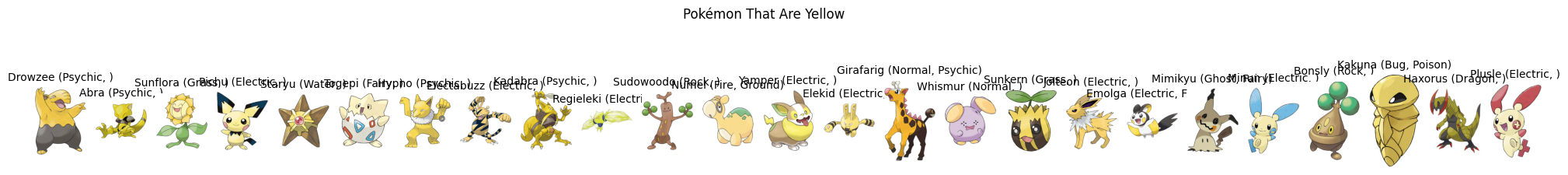}
                \includegraphics[width=\textwidth]{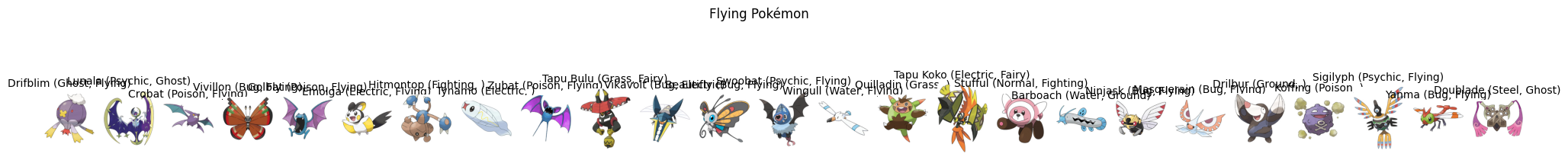}
    \caption{\textbf{An example of the Image Retrieval Task.} We highlight a prompt in this case: \texttt{"Blue Pokemon", "Pokemon That Look Like Birds", "Pokemon That Are Yellow", "Flying Pokemon"}, and we score the Precision@k, Recall @k, and f1 based on most relevant returns based on the query}
    \label{fig:enter-label}
\end{figure}

\end{document}